\documentclass[manuscript,review,anonymous]{amsart}
\usepackage{amsaddr}
\usepackage{tikz}
\usetikzlibrary{positioning,shapes.geometric}
\usepackage{multicol}
\usepackage{amsfonts}
\usepackage{amsthm}
\usepackage{amsmath}
\usepackage{amscd}
\usepackage[colorlinks=true, linkcolor=blue, urlcolor=blue, citecolor=blue]{hyperref}
\usepackage[utf8]{inputenc}
\usepackage{t1enc}
\usepackage[mathscr]{eucal}
\usepackage{indentfirst}
\usepackage{graphicx}
\usepackage{bbm}
\usepackage{graphics}
\usepackage{pict2e}
\usepackage{epic}
\usepackage{bbm}
\usepackage{lineno}
\makeatletter
\newcommand{\addresseshere}{%
  \enddoc@text\let\enddoc@text\relax
}
\makeatother

\numberwithin{equation}{section}
\usepackage[margin=1in]{geometry}
\usepackage[
backend=biber,
natbib,
style=authortitle,
citestyle=authoryear,
maxbibnames=99
]{biblatex}
\usepackage{epstopdf} 
\theoremstyle{plain}

\theoremstyle{definition}

\newtheorem{?}[Th]{problem}

\makeatletter
\renewcommand\subsection{\@startsection{subsection}{2}%
  \z@{.5\linespacing\@plus.7\linespacing}{-.5em}%
  {\normalfont\scshape}}
  
\def\specialsection{\@startsection{section}{1}%
  \z@{\linespacing\@plus\linespacing}{.5\linespacing}%
  {\normalfont}}
\def\section{\@startsection{section}{1}%
  \z@{.7\linespacing\@plus\linespacing}{.5\linespacing}%
  {\normalfont\scshape}}
\makeatother

\title{Loop Polarity Analysis to Avoid Underspecification in Deep Learning}
\author{Donald Martin, Jr.}
\address{Google Research}

\author{David Kinney}
\address{Washington University in St.\ Louis and Google Research}

\begin{document}
\setlength{\abovedisplayskip}{6pt}
\setlength{\belowdisplayskip}{6pt}
\setlength{\abovedisplayshortskip}{6pt}
\setlength{\belowdisplayshortskip}{6pt}

\maketitle

\begin{abstract}
Deep learning is a powerful set of techniques for detecting complex patterns in data. However, when the causal structure of that process is underspecified, deep learning models can be brittle, lacking robustness to shifts in the distribution of the data-generating process. In this paper, we turn to loop polarity analysis as a tool for specifying the causal structure of a data-generating process, in order to encode a more robust understanding of the relationship between system structure and system behavior within the deep learning pipeline. We use simulated epidemic data based on an SIR model to demonstrate how measuring the polarity of the different feedback loops that compose a system can lead to more robust inferences on the part of neural networks, improving the out-of-distribution performance of a deep learning model and infusing a system-dynamics-inspired approach into the machine learning development pipeline. 
\end{abstract}

\section{Introduction}
The use of neural networks to detect complex patterns in data (i.e., ``deep learning'') has emerged as one of the most powerful techniques for learning and inference in contemporary artificial intelligence (AI) and machine learning (ML) \citep{lecun2015deep}. However, limitations of the deep learning framework become apparent when it is applied to data that are generated by complex socio-technical systems characterized by feedback loops and temporal dynamics. Specifically, when deep learning is applied to data generated by complex causal systems, the models learned by neural networks are prone to brittleness or lack of robustness. Models are considered brittle when their predictive inference performance deteriorates upon receiving inputs - from real-world deployment domains - that fall outside of the distribution represented in the model training and evaluation data. This problem is also referred to as the “out-of-distribution (OOD) generalization problem” (\cite{shen2021towards}).\par

AI researchers have pointed to \textit{underspecification} in ML development pipelines as a root cause of poor OOD performance in deep learning models \citep{d2020underspecification}. Underspecification occurs in AI/ML development when a neural network is trained in an environment where there are multiple routes to optimal performance, because the underlying \textit{structure} of the problem domain is not appropriately specified. As D'Amour et al.\ put it:
\begin{quote}
    For example, the solution to an underdetermined system of linear equations (i.e.,
more unknowns than linearly independent equations) is underspecified, with an equivalence class
of solutions given by a linear subspace of the variables. In the context of ML, we say an ML
pipeline is underspecified if there are many distinct ways (e.g., different weight configurations) for
the model to achieve equivalent held-out performance on iid [i.e., independently and identically-distributed] data, even if the model specification and
training data are held constant. For example, the solution to an underdetermined system of linear equations (i.e.,
more unknowns than linearly independent equations) is underspecified, with an equivalence class
of solutions given by a linear subspace of the variables. In the context of ML, we say an ML
pipeline is underspecified if there are many distinct ways (e.g., different weight configurations) for
the model to achieve equivalent held-out performance on iid data, even if the model specification and
training data are held constant (\citeyear{d2020underspecification}, p. 3). 
\end{quote}
In cases where the relevant ML pipeline is underspecified, a neural network \textit{may} detect patterns in its training data that reflect the underlying structure of the data-generating process (DGP). However, it may also \textit{fail} to do so, instead learning spurious correlations that are not explained by any structural causal theory of the system under study (\cite{d2020underspecification,scholkopf2021towards}). In the latter case, deploying the model in novel, real-world environments can lead to poor performance. Indeed, there are now multiple real-world cases in which deep learning models' failures to encode important structural causal knowledge have caused societal harms in high-stakes domains (\cite{obermeyer2019dissecting,ensign2018runaway}).\par

In this paper, we turn to the system dynamics (SD) literature for strategies that will allow us to make progress on the underspecification problem in deep learning \citep{forrester1997industrial,sterman2010business}. A central tenet of system dynamics is that in order to understand a complex dynamical system, one must understand how the different feedback loops that compose the system drive overall system behavior. In particular, \citet{richardson1986dominant} has shown how one can gain a rich understanding of the dynamics of a system by grasping the relationship between the \textit{polarity} of a system's component feedback loops and the overall behavior of that system. We use this insight to show how neural networks that are trained on a data representation that measures the polarity of different feedback loops within an assumed data-generating process can achieve better OOD performance on crucial classification tasks than models trained on data sets that do not incorporate this same SD-inspired representation of features. Our techniques constitute an instance of ``manual feature engineering,'' in the sense that  we demonstrate how practitioners who are equipped with an SD-style understanding of the causal dynamics of a DGP can manually construct features that represent the polarity of the different feedback loops that constitute the DGP, which in turn enables them to build deep learning models that are robust to distribution shifts \citep{verdonck2021special}.\par

The remainder of this paper proceeds as follows. In Section \ref{sec:related}, we discuss previous work that is relevant to our discussion here, and situate our study within this context. In Section \ref{sec:nn}, we provide a basic overview of the elements of neural network architecture for classification that are relevant to our study. In Section \ref{sec:sd}, we introduce the loop polarity framework for understanding system behavior. In Section \ref{sec:cs}, we use the case study of a simple, simulated epidemic model to show how this polarity framework enables a manual feature engineering strategy that leads to more accurate classification on OOD data. We discuss the broader implications of this case study for the ML development pipeline in Section \ref{sec:discuss}, and conclude in Section \ref{sec:conc}.\par

\section{Related Work}\label{sec:related}
The work that most directly addresses the intersection of polarity frameworks and deep learning is found in \citet{schoenberg2020loops} and \citet{schoenberg2020building}. That work advocates the use of neural networks to build causal loop diagrams based on data with less structure. By contrast, in what follows we will assume that a particular causal loop structure, represented as a set of ordinary differential equations, generates a particular data set, and then exploit that assumption to represent the data in a way that allows for more effective neural network classification on OOD data. In this sense, our project is closer to work by \citet{kyono2021exploiting}, who show how training a neural network using a loss function that rewards a prediction’s coherence with a pre-assumed causal structure leads to improved inference. However, whereas that work assumes that data is generated by an acyclic causal structure, we assume an inherently cyclic causal structure for the data-generating process, represented mathematically by a system of ordinary differential equations. This allows us to exploit dynamical aspects of the data-generating process, such as polarity, that are not represented in an acyclic framework.

\citet{magliacane2018domain} show how knowledge of the causal structure of a DGP can enable better OOD inference about the data that would be produced under hypothetical interventions. Specifically, they provide an algorithm that uses causal knowledge to learn the features of data that are most predictive of system behavior under hypothetical interventions on variables. Here, we take a similar approach, learning coarse-grained features of data that are predictive of overall system behavior. However, by measuring the polarity of the quantities represented in our data, we are able to make inferences about the dynamics of the system (i.e., how the system is evolving over time).\par

Recent work by \citet{bongers2021foundations} and \citet{weinberger2020near}, has extended the structural causal modeling framework for generative inference to include cyclic relationships between random variables and differentiable stochastic processes. In this context, one can see our contribution here as partly a fusion of Kyono and van der Shaar’s approach to neural network inference and the cyclic nature of Bongers et al.’s generative models. However, Bongers et al.\ do not use their framework to measure structural parameters like polarity, nor do they explicitly apply their approach to the problem of data representation in the ML development pipeline. In both of these respects, our work expands on theirs by explicitly connecting a cyclic causal framework to both the dynamical systems literature and data representation.\par

In addition, recent work by \citet{geiger2023causal} generalizes the idea of “causal abstraction,” which considers how the variables of a causal model can be fused and coarse-grained to create simpler causal models (see \cite{chalupka2017causal,beckers2019abstracting,rubenstein2017causal}), to the cyclic (as opposed to acyclic) setting. They then use this more generalized notion of abstraction to coarse-grain the architectures of neural networks, to yield a more tractable, explainable architecture. This is related to our project here in that it exploits knowledge of cyclic causal structures to simplify and improve neural network inference. We also use the polarity framework to coarse-grain and simplify our representation of a data-generating process. However, our work is distinguished by the fact that we enable simpler, more tractable representations of the data that are used to train a neural network, rather than coarse-graining and abstracting the structure and parameter space of the neural network itself.\par

\section{The Neural Network Architecture}\label{sec:nn}
In contemporary deep learning, the most common technique for approximating the functional relationship between inputs and outputs to a DGP is to use a neural network. Though the formal details can get extremely complicated, at its core a neural network aims to approximate a function. Let $\mathcal{X}$ be some vector space representing inputs to that function, and $Y$ be some space of possible outcome of interest. We assume that there is a ``ground truth'' function $f:\mathcal{X}\rightarrow Y$ that generates data, in the form of output-input pairs in $\mathcal{X}\times Y$. To approximate this function, we begin by letting $l:Y\times Y\rightarrow \mathbbm{R}$ be a \textbf{loss function} such that $l(y,y_{obs})$ measures the distance between the guessed output $y$ and an observed output $y_{obs}$. The neural network aims to find a function $f^{*}$ that minimizes \textbf{expected risk} $R(f^{*})$, which is defined as follows:
\begin{equation}
    R(f^{*})=\int_{\mathcal{X}\times Y}l(f^{*}(\mathbf{x}),y)P(\mathbf{x},y),
\end{equation}
where $P$ is a data-generating probability distribution over $\mathcal{X}\times Y$. To find this function, the neural network is typically given \textbf{training data} consisting of pairs $\{(\mathbf{x}_{1},y_{1}),\dots,(\mathbf{x}_{n},y_{n})\}\sim P$. The neural network then aims to use this training data, through an algorithmic training procedure, to find a function that minimizes \textbf{expected empirical risk}, which is defined as follows:
\begin{equation}
    R_{emp}(f^{*})=\frac{1}{n}\sum_{i=1}^{n}l(f^{*}(\mathbf{x}_{i}),y_{i}).
\end{equation}
By aiming to find the function that minimizes this quantity, the neural network aims to find a function that captures as accurately as possible the true data-generating function $f$, provided that the training data are representative of all possible data sets that could be generated by $P$.

\begin{figure}
    \centering
    \includegraphics[scale=.5]{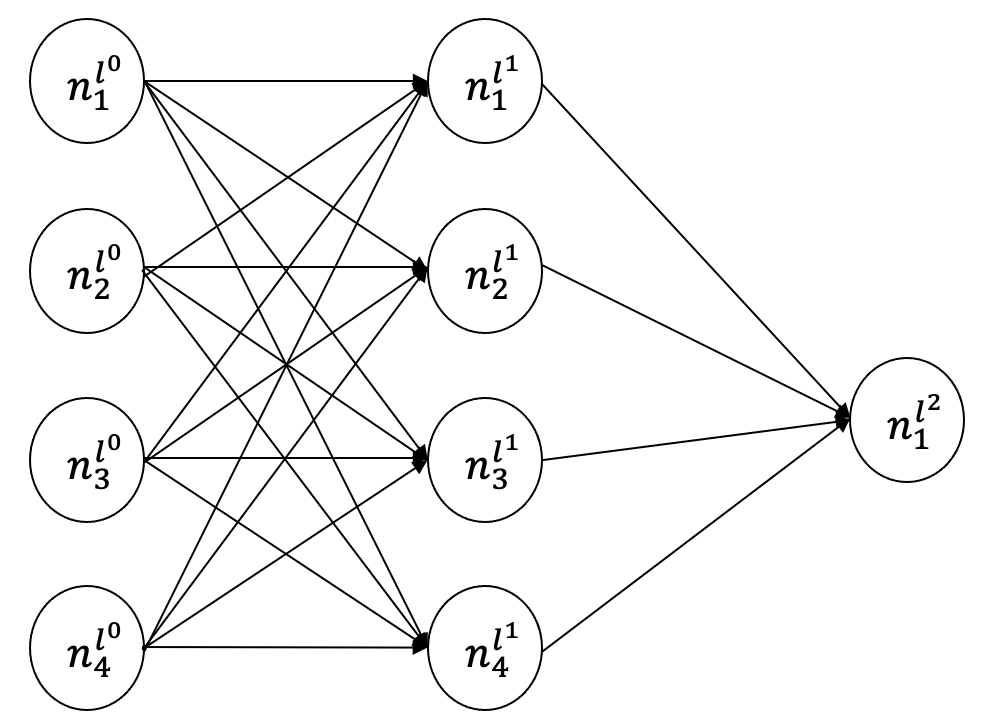}
    \caption{Simple neural network with a four-dimensional input layer, a single four-dimensional hidden layer, and one-dimensional output layer.}
    \label{fig:nn_arch}
\end{figure}

In practice, neural networks achieve this approximation of an often-complex target function by composing many simpler functions. To illustrate, consider a simple neural network called a \textbf{multilayer perceptron}. Such a network consists of layers of neurons. An initial input layer $l^{0}$ has the same dimensionality as each input vector $\mathbf{x}$. Subsequent intermediate or ``hidden'' layers $(l^{1},l^{2},\dots)$ may have more or less neurons than the input layer, and the final output layer consists of a single neuron. Each neuron $n_{i}^{l^{k}}$ in each hidden layer $l^{k}$ is a function whose value depends on the value of each neuron in the previous layer $(n^{l^{k-1}}_{1},\dots,n^{l^{k-1}}_{|l^{k-1}|})=l^{k-1}$, and a set of parameters $\theta_{n_{i}^{l^{k}}}^{n^{l^{k-1}}_{1}},\dots,\theta_{n_{i}^{l^{k}}}^{n^{l^{k-1}}_{|l^{k-1}|}}$. The final output layer $l^{K}$ contains a single neuron, which in the case of a binary classifier takes a value in the space $\{0,1\}$. We interpret the value of the neuron in the output layer as specifying whether or not the input vector belongs in a given class or not. See Figure~\ref{fig:nn_arch} for a representation of a simple multi-layer perceptron.\par

More complex architectures than the multilayer perceptron exist for learning a function from inputs to outputs in training data. For example, the ``long short-term memory'' (hereafter, ``LSTM'') neural network architecture has been shown to be especially adept at learning relationships in time series data \citep{hochreiter1996lstm}. This architecture is used in our computational experiments below to learn hidden parameter values for in-distribution data. While the details of this more complex architecture need not detain us here, the LSTM, like the multilayer perception, composes many different functions together to approximate the function that best represents the input-output relationships that partially characterize a data-generating process.

Since the value of each neuron in a neural network is determined both by other neurons in the network and a parameter vector, for a neural network to accurately approximate some function, the parameters of that function must be tuned so that the output of the neural network, given some input vector, tends to match the output of the target function, given the same vector. To tune parameters in this way, neural networks are fed training data consisting of input-output pairs $(\mathbf{x},y)$. Neural network training algorithms then aim to find parameters that enable accurate prediction of each output $y$ from the associated input $\mathbf{x}$, using any number of optimization procedures (e.g., gradient descent algorithms). The details of these training algorithms are ultimately tangential to our arguments here; the important point is that in order to learn parameters that enable accurate classification, a neural network must be provided training data.\par

What should be apparent from this exegesis is that very early in the deep learning analysis pipeline, practitioners have to make choices about the data that they use to train a neural network. The outputs of the same DGP can be represented in many different ways, and choices with regard to these data representations can make a difference to the success of training different neural networks. In what follows, we consider a method of data representation from the system dynamics literature (i.e., the ``polarity'' framework for representing the output of a data-generating process), and show how training a neural network on this data representation improves performance on data generated from the same causal structure, but with parameters sampled from very different distributions than those used to generate the parameters values for the model used to generate the training data. To our knowledge, ours is the first attempt to use loop-polarity data to train a neural network to classify the behavior of systems. By demonstrating good performance on OOD classification for neural networks trained on this data in a computational experiment, we invite further research on the part of the ML community aimed at using similar data to train predictive algorithms in a variety of applications.\par

While neural network training algorithms have access to training data, they do not have access to the underlying causal or dynamical structure of the DGP. This leads to an underspecification problem in which a training algorithm may converge on what appears to be an optimal representation of the DGP, but which lacks the fidelity to the data-generating structure that is needed to generalize results to other domains in which accidental features of the DGP may have changed. As noted above, this a general challenge for the application of deep learning to critical problem domains. In what follows, we investigate how insights from system dynamics can be used to address and mitigate these kinds of underspecification concerns.\par

\section{Feedback Loop Polarity for System Understanding}\label{sec:sd}
We turn now to the polarity framework for understanding system behavior. Our approach here takes its inspiration from contributions to the system dynamics literature due to \citet{richardson1986dominant} and \citet{hayward2014model}. Within this framework, different possible behaviors of a system over time are represented by stochastic processes of the form $X:T\times\Omega\rightarrow \mathbbm{R}$, which are called \textbf{levels}. We assume here that levels are at least twice differentiable. This level can then be represented as part of a \textbf{feedback loop} $(X,\dot{X}=\frac{\partial X}{\partial t})$ consisting of the level and its \textbf{inflow rate}, or rate of change over time. The \textbf{polarity} of a feedback loop is then given by $\texttt{sign}(\frac{\partial\dot{X}}{\partial X})$, i.e., the sign of the rate of change in the rate of change in $X$ as $X$ changes. Thus, if the rate of change of a level increases as the overall measure of that behavior increases, then the corresponding feedback loop is said to have positive polarity. If the rate decreases as the overall measure increases, then the loop is said to have negative polarity. The inflow rate of a given level can be represented as determined by the inflow rates of other levels within the system, so that for a given level $X_{0}$, $\frac{\partial X_{0}}{\partial t}=\varphi(X_{1},\dots,X_{n})$, where $\varphi$ is a function and $\{X_{1},\dots,X_{n}\}$ is a set of other levels within the system. This is another way of saying that the dynamics of the system can be represented using a set of differential equations.\par

\begin{figure}
    \centering
    \includegraphics[scale=.5]{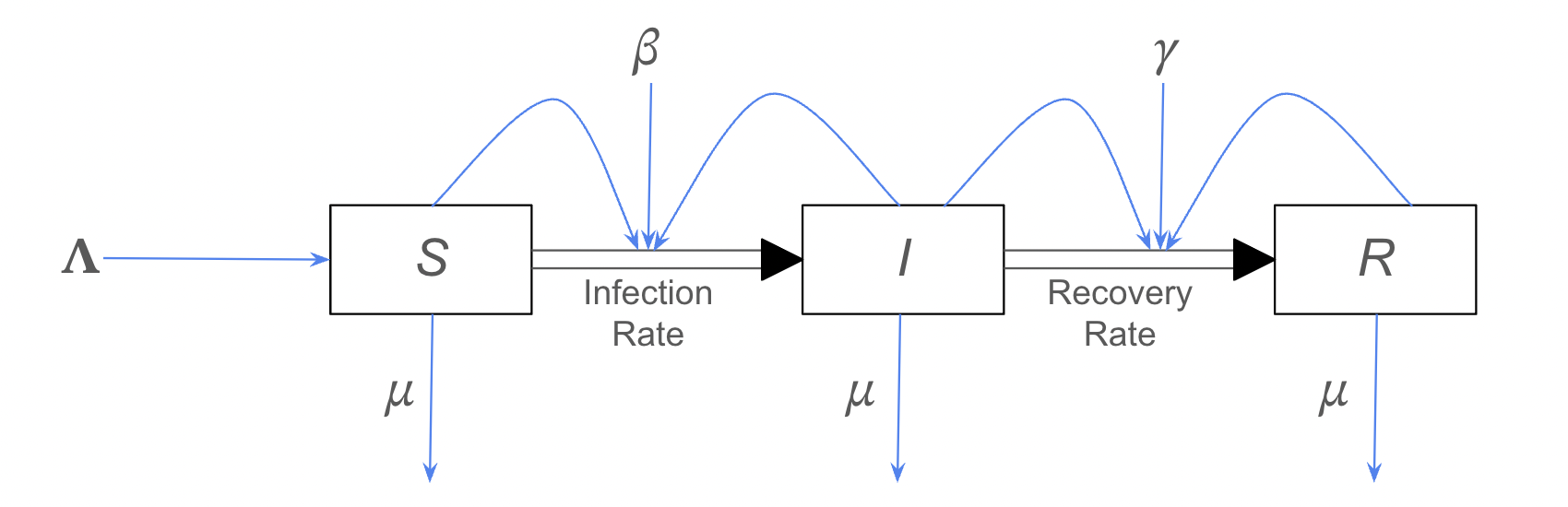}
    \caption{Stock-flow diagram representing the set of differential equations that define the epidemic system considered here. Note that $\Lambda$ is the birth rate in the population, $\mu$ is the death rate, $\beta$ is the average number of interactions between infected and susceptible people in the population per time step, and $\gamma$ is the recovery rate.}
    \label{fig:stockflow}
\end{figure}

One key insight of the dynamical systems literature is that in many instances, the polarities of the component feedback loops of a data-generating process tend to determine the overall behavior of the system composed of those functions. Indeed, there is a sense in which understanding this relationship between sub-system behavior and overall system behavior is just what it means to understand \textit{how a system works}. Following Hayward and Boswell (\citeyear{hayward2014model}, p.\ 41-3), we illustrate the polarity framework using the case study of a simple SIR (Susceptible, Infected, Recovered) model of an epidemic with vital dynamics (\cite{kermack1927contribution}). Note that our choice of model here is for illustrative purposes, and that one could explore applying our approach on other epidemic models, such as the Susceptible-Infected-Susceptible (SIS) model. The model consists of five parameters and three first-order differential equations, which are as follows:
\begin{align}
    \Lambda(t) &:= \text{birth rate}\label{eq:sirmodel1}\\
    \mu(t) &:= \text{death rate}\label{eq:sirmodel2}\\
    \gamma(t) &:= \text{recovery rate}\label{eq:sirmodel3}\\
    \beta(t) &:= \text{average number of interactions with other people per time-step}\label{eq:sirmodel4}\\
    N &:= \text{population}\label{eq:sirmodel5}\\
    \dot{S}(t) &= \Lambda(t)N - \mu(t) S(t) - \frac{\beta(t) I(t)S(t)}{N}\label{eq:sirmodel6}\\
    \dot{I}(t) &= \frac{\beta(t) I(t)S(t)}{N} - \gamma(t) I(t) - \mu(t) I(t)\label{eq:sirmodel7}\\
    \dot{R}(t) &= \gamma(t) I(t) - \mu(t) R(t)\label{eq:sirmodel8}
\end{align}
Apart from the population $N$, which does not vary with time, all other parameters and functions herein are levels, as they are stochastic processes representing quantities of interest in the model. At any given time $t$ in the course of an epidemic, the rates of change in the number of susceptible, infected and recovered members of the population is determined by the solution to these differential equations. At a qualitative level, we can see that the rate of change in the number of susceptible people is a function of the birth rate, the number of susceptible people who die during a given time step, and the number of interactions between infected and susceptible people as a proportion of the total population. The rate of change in the number of infected people is a function of the number of interactions between infected and susceptible people as a proportion of the total population, the number of infected people who recover, and the number of infected people who die. The rate of change in the number of recovered people is a function of the number of infected people who recover and the number of recovered people who die. See Figure~\ref{fig:stockflow} for a stock-flow diagram representing the system of differential equations that defines the epidemic.\par

In keeping with the polarity approach to understanding system behavior, the data-generating process of the epidemic can be decomposed into each summand of the governing ODEs shown in Eqs.\ \ref{eq:sirmodel6}-\ref{eq:sirmodel8} above. The integral of each summand up to a given time step can then be represented as part of its own loop. These loops are represented as follows:
\begin{align}
    \mathcal{L}_{1} &:= \left(\int_{0}^{t}\Lambda(t^{\prime})N\text{d}t^{\prime},\Lambda(t)N\right) \ \text{(the birth rate)}\label{eq:firstloop}\\
    \mathcal{L}_{2} &:= \left(\int_{0}^{t}\mu(t^{\prime}) S(t^{\prime})\text{d}t^{\prime},\mu(t) S(t)\right)\ \text{(the number of susceptible people who die)}\\
    \mathcal{L}_{3} &:= \left(\int_{0}^{t}\frac{\beta(t^{\prime}) I(t^{\prime})S(t^{\prime})}{N}\text{d}t^{\prime},\frac{\beta(t) I(t)S(t)}{N}\right) \ \text{(infected-susceptible interactions over $N$)}\\
    \mathcal{L}_{4} &:= \left(\int_{0}^{t}\gamma(t^{\prime}) I(t^{\prime})\text{d}t^{\prime},\gamma(t) I(t)\right) \ \text{(the number of infected people who recover)} \\
    \mathcal{L}_{5} &:= \left(\int_{0}^{t}\mu(t^{\prime}) I(t^{\prime})\text{d}t^{\prime},\mu(t) I(t)\right) \ \text{(the number of infected people who die)}\\
    \mathcal{L}_{6} &:= \left(\int_{0}^{t}\mu(t^{\prime}) I(t^{\prime})\text{d}t^{\prime},\mu(t) I(t)\right) \ \text{(the number of recovered people who die)}\label{eq:lastloop}
\end{align}
Each of these loops represents a component feedback loop of the overall DGP for this hypothetical epidemic. As described above, a hallmark of the system dynamics literature is the attempt to understand how the behavior of these component feedback loops drives overall system behavior. In many contexts, a crucial measure of of overall system behavior is the polarity of the loop $\left(I(t),\frac{dI(t)}{dt}\right)$ representing the overall level of infections at a given time step (\cite{center_for_disease_control_2021}). This quantity tells us whether, as infections increase, the \textit{rate of change in infections is also increasing} (indicating exponential growth and an urgent need to martial resources), or whether, as infections increase, the \textit{rate of change in infections is decreasing}, indicating that the epidemic is nearing its peak. For instance, if the total number of infected people that have recovered from the epidemic is plateauing, then this binary judgment may be a powerful tool for estimating the overall polarity of the number of infections. In the SIR model presented here, this system-level polarity measure may be driven by the polarity of the component loops listed above. We hypothesize that if we can understand this relationship between the polarity of the component feedback loops and the overall behavior of the system, then we can understand the system driving the epidemic in a way that avoids underspecification and if more robust to accidental changes in the data-generating process.\par

In what follows, we test this hypothesis in a computational experiment. We begin by simulating data from an epidemic that is generated by the SIR model defined above, and show how we are able to use deep learning to accurately infer the unobservable parameters of that epidemic system (i.e., the birth rate, death rate, recovery rate, and average number of interactions).  We then show that we can gain an understanding of the underlying DGP for an epidemic by using the parameters we have inferred in-distribution to re-simulate the epidemic, and learn the relationships between the component loops of the system and overall system behavior - relationships that are robust to distributional shifts.\par

\section{Demonstration on a Simple Epidemic Model}\label{sec:cs}
We begin our computational experiment by simulating data from a hypothetical epidemic.\footnote{Python code for reproducing the computational experiments conducted in this section is included with this submission as supplemental material.} We use the \texttt{ODEInt} Python package for generating simulated data based on a set of differential equations, generating time-series data recording the number of susceptible, infected, and recovered people in a population based on the equations shown in Eqs.~\ref{eq:sirmodel6}-\ref{eq:sirmodel8}. Moreover, we suppose that at each time step in an epidemic, the \textit{un}observed parameters $\Lambda(t)$, $\mu(t)$, $\gamma(t)$, and $\beta(t)$ are generated by sampling from the following distributions:
\begin{align}
    \Lambda(t) &\sim \text{Beta}(2,\frac{2-.0002}{.0001}) \ \text{(i.e., a Beta distribution with a mean of $.0001$)}\label{eq:indistfirst}\\
    \mu(t) &\sim \text{Beta}(2,\frac{2-.0002}{.0001}) \ \text{(i.e., a Beta distribution with a mean of $.0001$)}\\
    \gamma(t) &\sim \text{Beta}(2,\frac{2-.2}{.1}) \ \text{(i.e., a Beta distribution with a mean of $.1$)}\\
    \beta(t) &\sim \text{Poisson}(5) \ \text{(i.e., a Poisson distribution with a mean of $5$)}\label{eq:indistlast}
\end{align}
Using this simulation process, we collect 10,000 synthetic data samples:\ 100 epidemics generated via the same process, each with 100 time steps. The observable state of the epidemic at each time step is a real-valued, three-entry vector representing the number of susceptible, infected, and recovered people in the population at each time step.\par 

Next, we use an LSTM-based neural network architecture to learn a mapping from observable states of the epidemic at each time step to the unobserved features $\Lambda(t)$, $\mu(t)$, $\gamma(t)$, and $\beta(t)$ at the same time step. To do this, we suppose that we do in fact have access to the values of these unobserved parameters, and save the value for each unobserved parameter generated at each time step during the simulation described in the previous paragraph. This provides a training data set for learning a function mapping each observed state of the epidemic at a given time step to the set of unobserved parameters that sampled at that same time step. Once we obtain this function, we are able to re-simulate the epidemics, obtaining our unobserved parameters not by sampling from a distribution, but by obtaining the output of the function learned by our LSTM-based neural network when given as input the observable state of the epidemic at each time step during the second simulation. This process is meant to mimic a real-world process in which we have a previously obtained training data set matching observed to unobserved states of the epidemic, which we then use to train an LSTM-based neural network to learn a mapping from observed to unobserved states in some new epidemic which we know to be generated by the same process. Once this mapping is learned, we can use it to \textit{infer} the behavior of a new epidemic where we know that data is generated by the same process, including the sampling of parameters from the same distribution. In our simulations, we find in-distribution parameter estimation using deep learning to be highly accurate, with a mean-weighted variance between predicted and actual values of $6.2\times10^{-5}$ for estimates of $\Lambda(t)$, $6.5\times10^{-5}$ for estimates of $\mu(t)$, $0.1$ for estimates of $\gamma(t)$, and $0.9$ for estimates of $\beta(t)$. We can then use our inferred unobserved parameters to simulate a new epidemic.\par

When we simulate a new epidemic using inferred parameters, we can take ourselves to have access to the unobserved parameters generating our synthetic data. Thus, we are able to study, in a synthetic setting, the relationship between the polarity of the sub-loops defined in Eqs.~\ref{eq:firstloop}-\ref{eq:lastloop} above and the overall polarity of the level of infections in the epidemic. Specifically, for each three-step time interval in each of our simulated epidemics, we manually engineer a six-entry feature vector:
\begin{equation}
\mathbf{x}^{P}=\left[\texttt{polarity}\left(\mathcal{L}_{1}\right),\dots,\texttt{polarity}\left(\mathcal{L}_{6}\right)\right],
\end{equation}
where for each loop $\mathcal{L}_{i}$,
\begin{equation}
    \texttt{polarity}\Big(l_{i},\frac{dl_{i}}{dt}\Big)=\texttt{sign}\Big(\frac{[l_{i}(t+2)-l_{i}(t+1)]-[l_{i}(t+1)-l_{i}(t)]}{[l_{i}(t+2)-l_{i}(t)]}\Big). 
\end{equation}
Similarly, the polarity of the overall level of infections in the epidemic is defined as follows:
\begin{equation}
    \texttt{polarity}\Big(I,\frac{dI}{dt}\Big)=\texttt{sign}\Big(\frac{[I(t+2)-I(t+1)]-[I(t+1)-I(t)]}{[I(t+2)-I(t)]}\Big). 
\end{equation}
Using the data from our re-simulated epidemics, we train a multi-layer perception to classify the overall system behavior $\texttt{polarity}\Big(I,\frac{dI}{dt}\Big)$ during a three-time-step interval, where the input to the classifier function is the vector $\mathbf{x}^{P}$ representing the polarity of the component loops of the system over the same interval. Recall that learning this kind of relationship between component loop behavior and overall system behavior is seen within the SD literature as central to gaining a genuine understanding of that system. By manually engineering the polarity features $\mathbf{x}^{P}$, we enable a neural network to encode exactly this kind of systematic understanding of the overall data-generating process.\par

Next, we show that, in keeping with our hypothesis, a polarity-based understanding of system behavior encoded in the mapping learned by a neural network can improve the out-of-distribution performance of that mapping. To test this, we generate OOD data by first sampling the following meta-parameters:
\begin{align}
    \bar{\Lambda} &\sim \text{Beta}(2,\frac{2-.02}{.01})\label{eq:oodfirst}\\
    \bar{\mu} &\sim \text{Beta}(2,\frac{2-.02}{.01})\\
    \bar{\gamma} &\sim \text{Beta}(2,\frac{2-.02}{.01})\\
    \bar{\beta} &\sim \text{Poisson}(15)
\end{align}
We then use each of these meta-parameters to generate 100 epidemics, each with 100 time steps, where at each time step parameters are generated by sampling from the following distributions:
\begin{align}
    \hat{\Lambda}(t) &\sim \text{Beta}(2,\frac{2-2\bar{\Lambda}}{\bar{\Lambda}}) \ \text{(i.e., a Beta distribution with a mean of $\bar{\Lambda}$)}\\
    \hat{\mu}(t) &\sim \text{Beta}(2,\frac{2-2\bar{\mu}}{\bar{\mu}}) \ \text{(i.e., a Beta distribution with a mean of $\bar{\mu}$)}\\
    \hat{\gamma}(t) &\sim \text{Beta}(2,\frac{2-2\bar{\gamma}}{\bar{\gamma}}) \ \text{(i.e., a Beta distribution with a mean of $\bar{\gamma}$)}\\
    \hat{\beta}(t) &\sim \text{Poisson}(\bar{\beta}) \ \text{(i.e., a Poisson distribution with a mean of $\bar{\beta}$)\label{eq:oodlast}}
\end{align}   
This sampling procedure is very likely to result in epidemics generated by sampling from distributions with very different means from those used to generate the training data. Nevertheless, if we give ourselves access to the unobserved parameters, and use them to obtain the polarity features $\mathbf{x}^{P}$ for the out-of-distribution data, we find that our neural network that was trained on a data generated in a different distributional setting \textbf{is still able to predict the polarity of the overall level of infections from the polarity of the component sub-loops of the epidemic with 74.3\% accuracy} across 20 iterations of 100 simulated 100 time-step epidemics, an outcome with a probability of occurring through random guessing that is less than $.0001$. We take this result to demonstrate how manually engineered features inspired by system dynamics can reduce underspecification and improve the out-of-distribution performance of deep learning.\par

To underscore this point about the ability of SD-inspired approaches to improve the OOD performance of deep learning, we conduct a second computational experiment to compare the performance of our SD-inspired classifier with a more straightforward approach that simply uses the observed data from a epidemic to classify the polarity of the level of infections in a system. We begin by once again generating 100 simulated epidemics, each with 100 time steps, sampling parameters using the distributions shown in Eqs.~\ref{eq:indistfirst}-\ref{eq:indistlast}. Next, we train two different multi-layer perceptrons to classify the polarity of the level of infections in a system over a three-step time interval. The first uses aims to classify the polarity of the level of infections using the information contained in the polarity vector $\mathbf{x}^{P}$ defined above. The second uses a nine-entry ``raw data'' vector $\mathbf{x}^{RD}$ that contains the number of susceptible, infected, and recovered members of the population at each time step in a three-step interval.\par

\begin{figure}
    \centering
    \includegraphics[scale=.25]{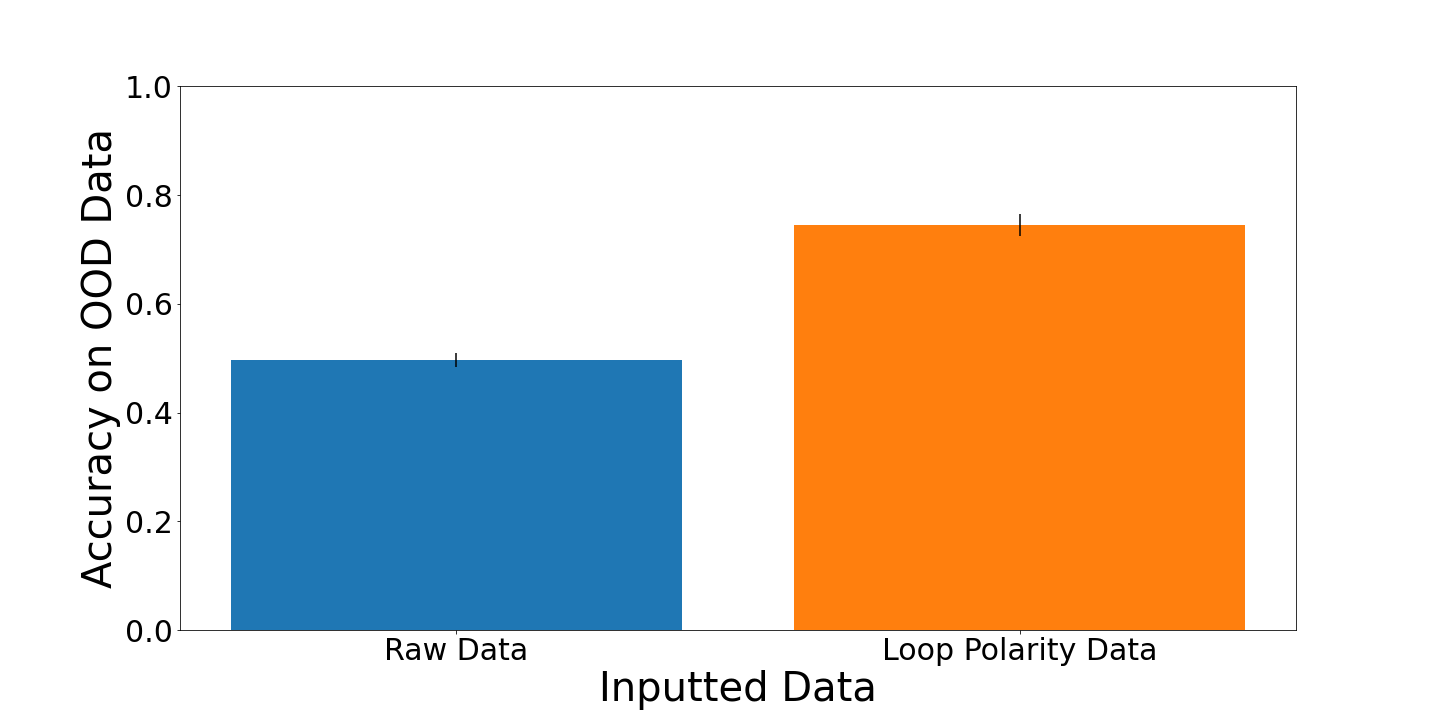}
    \caption{Mean accuracy across 20 iterations of 100 simulated OOD epidemics of both a neural network trained on raw data and a neural network trained on polarity data.}
    \label{fig:ood_comp}
\end{figure}

Next, we compare the performance of a multi-layer perceptron trained on the in-distribution data $\mathbf{x}^{P}$ and a multi-layer perceptron trained on the in-distribution data $\mathbf{x}^{RD}$ when we move to the out-of-distribution setting in which data is generated by the distributions shown in Eqs.~\ref{eq:oodfirst}-\ref{eq:oodlast}. Figure~\ref{fig:ood_comp} shows the accuracy of each multi-layer perceptron across 20 iterations of 100 simulated out-of-distribution epidemics. The model trained on the raw data achieves an average accuracy of $.497$ across all pandemics, performing slightly better than chance given that approximately $48.4\%$ of time steps intervals have a positive-polarity loop for the level of infections. By contrast, the model trained on the polarity data achieves an average accuracy of $.745$, performing much better than chance. This difference in means is highly statistically significant ($t=200.69$, $p<.001$). Thus, a neural network that takes as input data representing the polarity of loops containing various summands in the data-generating process is able to significantly outperform a neural network trained on the raw data produced by the same data-generating process, once each neural network is asked to make predictions on data that is not in its training set, and which is generated via a process with the same structure, but different statistical distributions over key parameters. We note that the performance of both neural networks on the OOD data is closely tied to their training accuracy; training accuracy is much greater for the model trained on polarity data as compared to the model trained on raw data. This suggests that the improved performance is due largely to the patterns in the loop polarity data being more easily learned than the patterns in the raw data. This is striking, given that the loop polarity of the rate of infections can be calculated analytically from raw data, but not from the loop polarity data.\par

\section{Discussion}\label{sec:discuss}
Our findings from this computational experiment have broader implications for the optimal functioning of the machine learning development pipeline. At a high level of abstraction, a typical workflow in developing ML applications involves:\ 1) observing data from a system, 2) devising a mathematical representation of that data, 3) training a neural network model on that data, 4) using that data to make OOD classifications. Our results above highlight the importance of data representation in the success of this pipeline. Choices we make about how data is represented as it is measured and collected influence the training of neural network classifiers, which in turn influence the success of these models in OOD testing scenarios. Importantly, implementing the polarity scheme for representing data \textit{requires domain knowledge of the kind that is emphasized in the practice of system dynamics}. Namely, one must know the summands of the ODEs that represent the dynamics of the system, and know that measuring their polarity is key to inferring the overall polarity of the number on infected people. This is a perspective on the data that requires some background in epidemiology and the modeling of dynamical systems; unlike the raw data representation, it cannot be measured from an entirely naive perspective on the nature of the data-generating process.\par 

\begin{figure}
    \centering
    \includegraphics[scale=.25]{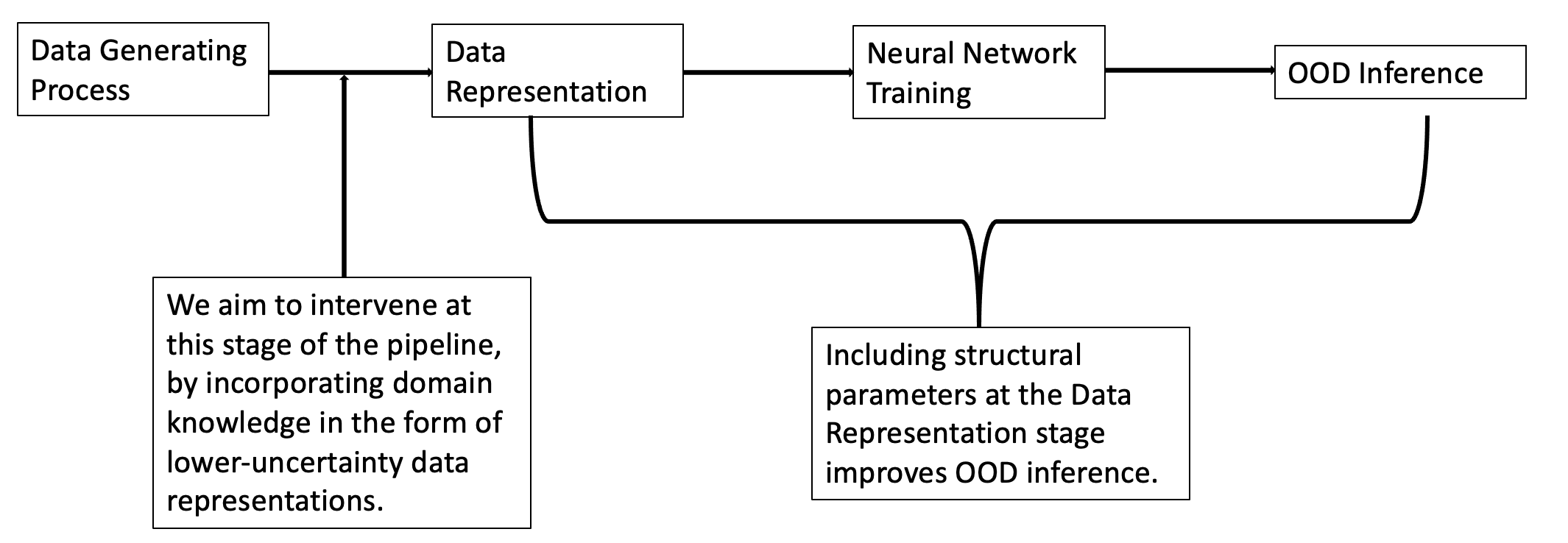}
    \caption{A high-level overview of the ML development pipeline and our proposed intervention.}
    \label{fig:pipeline}
\end{figure}

The perspective on the nature of dynamical systems that recommends decomposing those systems into summands and measuring the polarity of those summands is not one that is well-represented in existing machine learning practice. The results here indicate that in at least some contexts, this perspective \textit{should} be represented, as it can enable data representations that allow for more accurate classification of OOD data. To this end, we recommend that, throughout the development of the ML/AI pipeline, ML and AI developers should draw on the literature in system dynamics. They should also receive input - in the form of causal theories - from system dynamics practitioners, problem domain experts, and stakeholders. This is especially true in the data representation stage of that pipeline, where the emphasis is not on designing optimal architectures for function approximation, but is instead on how to interpret and optimally represent the outputs of a data-generating process. It is at this earlier (but still deeply important) stage that domain-specific causal theories with reduced epistemic uncertainty about the nature of the data-generating can be leveraged to enable better OOD performance. Figure~\ref{fig:pipeline} provides a schematic representation of our normative recommendation.\par

\section{Conclusion}\label{sec:conc}
There are several avenues for future work that build on our findings here. First, we note that the polarity framework is just one example of the many ways that experts in system dynamics have formalized the concept of understanding system behavior. In future work, we aim to explore whether other SD-inspired methods can show similar fecundity in reducing underspecification in deep learning. Second, we plan to extend our results beyond the synthetic setting, demonstrating similar performance for the results presented here on real-world data sets. Finally, while we have demonstrated the feasibility of our approach in a specific case study, we hope in the near future to both extend our approach to other dynamical systems beyond the SIR model, and to provide a theoretical guarantee or closed-form proof of the general ability of the polarity framework to improve out-of-distribution classification of system behavior.\par

\printbibliography
\end{document}